\documentclass{article} %
\usepackage{iclr2025_conference,times}

\usepackage{amsmath,amsfonts,bm}

\def\eqref#1{equation~\ref{#1}}

\def\1{\bm{1}}

\DeclareMathAlphabet{\mathsfit}{\encodingdefault}{\sfdefault}{m}{sl}
\SetMathAlphabet{\mathsfit}{bold}{\encodingdefault}{\sfdefault}{bx}{n}

\usepackage{hyperref}
\usepackage{url}

\usepackage{array}
\usepackage{colortbl}
\usepackage{adjustbox}
\usepackage{booktabs}
\usepackage{multirow}
\usepackage{paralist}

\usepackage{graphicx}
\usepackage{caption}
\usepackage{subcaption}
\usepackage{adjustbox}
\usepackage{pifont}
\usepackage{float}

\usepackage{color}
\usepackage{xcolor}
\usepackage{xspace}
\usepackage{enumitem}
\usepackage[most]{tcolorbox}
\usepackage{tabularray}

\newcommand{\cmark}{\ding{51}}

\title{Natural Language Inference Improves \\Compositionality in Vision-Language Models}

\author{Paola Cascante-Bonilla$^{1,2}$ %
\, Yu Hou$^{1}$ \, Yang Trista Cao$^{3}$ \, Hal Daumé III$^{1}$ \, Rachel Rudinger$^{1}$ \\
$^{1}$University of Maryland, College Park \, $^{2}$Stony Brook University \, $^{3}$University of Texas at Austin
}

\iclrfinalcopy %

\definecolor{bggray}{rgb}{0.95, 0.95, 0.95}
\usepackage[%
    framemethod=tikz,
    skipbelow=\topskip,
    skipabove=\topskip
]{mdframed}
\mdfsetup{%
    leftmargin=0pt,
    rightmargin=0pt,
    backgroundcolor=bggray,
    middlelinecolor=black,
    roundcorner=3
}
\newtcolorbox[list inside=prompt,auto counter,number within=section]{prompt}[1][]{
    colbacktitle=black!60,
    fonttitle=\small,
    coltitle=white,
    fontupper=\footnotesize,
    boxsep=4pt,
    left=0pt,
    right=0pt,
    top=0pt,
    bottom=0pt,
    boxrule=1pt,
    #1,
}

\NewDocumentCommand{\paola}{ mO{} }{\noindent\textcolor{purple}{\textsf{\small[#1]}\textsuperscript{\textit{Paola}}}}

\begin{document}

\maketitle

\begin{abstract}
    Compositional reasoning in Vision-Language Models (VLMs) remains challenging as these models often struggle to relate objects, attributes, and spatial relationships. %
Recent methods aim to address these limitations by relying on the semantics of the textual description, using Large Language Models (LLMs) to break them down into subsets of questions and answers. 
However, these methods primarily operate on the surface level, failing to incorporate deeper lexical understanding while introducing incorrect assumptions generated by the LLM. 
In response to these issues, we present Caption Expansion with Contradictions and Entailments (\textsc{Cece}), a principled approach that leverages Natural Language Inference (NLI) to generate entailments and contradictions from a given premise. 
\textsc{Cece} produces lexically diverse sentences while maintaining their core meaning. 
Through extensive experiments, we show that \textsc{Cece} enhances interpretability and reduces overreliance on biased or superficial features. 
By balancing \textsc{Cece} along the original premise, we achieve significant improvements over previous methods without requiring additional fine-tuning, producing state-of-the-art results on benchmarks that score agreement with human judgments for image-text alignment, and achieving an increase in performance on Winoground of $+19.2\%$ (group score) and $+12.9\%$ on EqBen (group score) over the best prior work (finetuned with targeted data). 
Project page: \textcolor{blue}{\url{https://cece-vlm.github.io/}}

\end{abstract}

\section{Introduction}

Trained with internet-scale data, large-scale Vision-Language Models (VLMs) often struggle to relate objects, attributes, understand spatial relationships, and grasp subtle changes in meaning due to small variations in images or word order~\citep{thrush_and_ross2022winoground, winogroundhard, wang2023equivariant, bitton2023breaking, yuksekgonul2023and, tong2024eyes, saxon2024evaluates, fu2024blink}. 
With a seeming inability to handle semantically modular scenarios, their opaque nature makes it difficult to understand their decision-making processes~\citep{dziri2023faith, kamath2023s, kamath2024the}. 
Furthermore, internal biases play a major role in affecting the model’s performance across various tasks
~\citep{zhou2022vlstereoset, tiong-etal-2024-measuring, howard2024socialcounterfactuals, fraser-kiritchenko-2024-examining, raj2024biasdora}.
Recent works have explored ways to mitigate these issues by breaking down a problem into smaller tasks.
Typically, a Large Language Model (LLM) is prompted to create small programs (i.e., Visual Programming (VP)~\citep{gupta2023visual, hu2023tifa, suris2023vipergpt, subramanian-etal-2023-modular, cho2024visual,  2024proptest, hu2024visual}) or deconstruct the textual description in a set of validation questions with their corresponding expected answers (i.e., Sentence Decomposition via Semantics (SDS)~\citep{cho2023davidsonian, wu2023rolechainofthoughtcomplexvisionlanguage, yarom2024you, mitra2024compositional,  zhang2024cocotcontrastivechainofthoughtprompting, Wan2024CRG}). 
While these methods provide interpretability, they also tend to degrade the VLM performance when evaluating challenging benchmarks that introduce pairs of images and captions that require extensive real-world knowledge and reasoning~\citep{lin2024vqascore}.

\begin{figure}[t]
    \centering
    \vspace{-0.5cm}
    \begin{subfigure}{\textwidth}
        \centering
        \includegraphics[width=\textwidth]{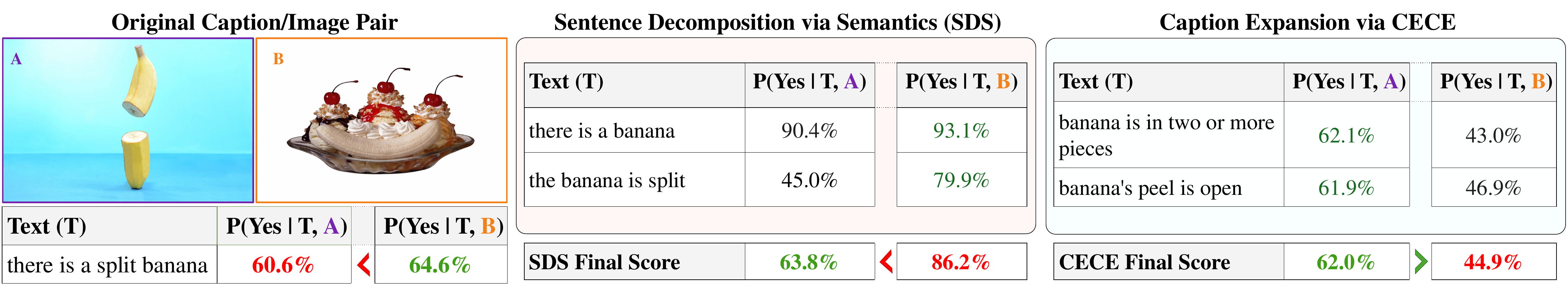} 
    \end{subfigure}
    \vspace{-0.15cm}
    \begin{subfigure}{\textwidth}
        \centering
        \includegraphics[width=\textwidth]{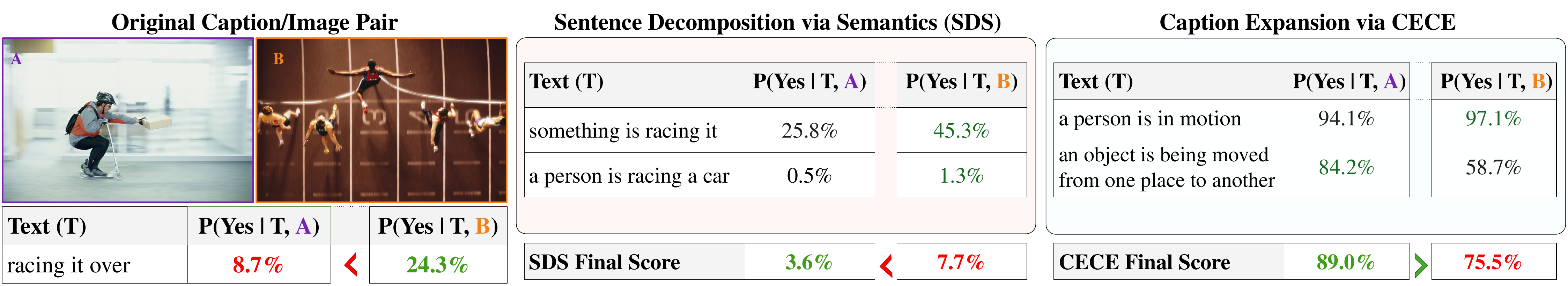} 
    \end{subfigure}
    \caption{Examples from Winoground dataset. The first column shows the output of LLaVa-1.6 when computing the likelihood of answering \textit{``yes”} given the image and text. The second column shows the sentence decomposition proposed in prior work (SDS), which follows the original caption semantics. The third column shows our proposed Caption Expansion with Contradictions and Entailments (\textsc{Cece}). In all cases, the model is only allowed to evaluate one image and text at a time.}
    \label{fig:figure1}
    \vspace{-0.5cm}
\end{figure}

Consider Figure~\ref{fig:figure1}; the first block shows two images with bananas. If we use a VLM to compute the likelihood of answering \textit{“yes”} given each image and the text \textit{“there is a banana split”}, the VLM incorrectly assigns a higher probability to the second image.
When decomposing the sentence through SDS,
the LLM will output sentences like: \textit{“there is a banana”} and \textit{``the banana is split"} (output examples taken from~\cite{cho2023davidsonian}). 
Although the SDS decomposition seems correct, it preserves the same lexical surface of the text and is unable to incorporate additional information that could be leveraged by the VLM to make a correct prediction -- both images contain bananas, but the VLM might have learned a stronger correlation of a \textit{banana split} being a \textit{dessert}. 
Likewise, if we look at the second block in Figure~\ref{fig:figure1} and follow the same process, the SDS output will not only preserve the same lexical surface, but will make wrong assumptions and introduce some of the biases present in the LLM (e.g., \textit{racing} might have strong correlations with \textit{cars}). 

To address these limitations, we introduce \textbf{Caption
Expansion with Contradictions and Entailments (\textsc{Cece})}, a principled approach that leverages Natural Language Inference (NLI) to improve the compositional capabilities of VLMs. 
With \textsc{Cece}, we instruct
an LLM to use NLI -- which is used to determine the relationship between two sentences, a premise and a hypothesis~\citep{bowman-etal-2015-large} --, and generate entailments (hypotheses that logically follow from the premise) and contradictions (hypotheses that are logically incompatible with the premise). In this way, the LLM is instructed to produce lexically diverse sentences while preserving the underlying meaning of the original captions. 
It is important to note that while the outputs generated by SDS can be considered a subset of entailments, \textsc{Cece} expands the scope by generating a wider variety of captions that capture nuanced relationships beyond what SDS would typically output, including both entailments and contradictions.
This allows the LLM to leverage common sense and break down the captions using world knowledge, while also mitigating the hallucinations and biases introduced by the LLM.

Revisiting Figure~\ref{fig:figure1}, \textsc{Cece} generates sentences that are lexically different from the original caption but preserve its meaning
(e.g., \textit{``split"} entails dividing things \textit{``in one or more pieces``}). 
By expanding the captions with both entailments and contradictions, the LLM incorporates world knowledge and common sense while mitigating hallucinations and biases. For example, \textit{``racing it over"} is parsed into \textit{``a person in motion"} and \textit{``an object being moved from one place to another"}, which entails a different perceptual inference of the caption, and reduces the likelihood of stereotypical associations. %
As a whole, \textsc{Cece} goes beyond the lexical boundaries of SDS, introducing richer contextual information for improved reasoning.

To leverage both entailments and contradictions generated via \textsc{Cece}, we evaluate the likelihood of the VLM answering \textit{``yes"} given the image and each entailment, and the likelihood of the VLM answering \textit{``no"} given the image and each contradiction. These scores are then aggregated using a weighting value to balance the contributions of both entailments and contradictions. 
With \textsc{Cece}, we incorporate both positive and negative reasoning cues: entailments provide semantic inclusion (focusing on subset relations), while contradictions provide semantic exclusion (focusing on subset complements).
In addition, we aggregate the VLM likelihood of answering \textit{``yes"} given the image and original caption.
We found that including the original caption provides additional context for balancing out the VLM outputs. In a way, the original caption serves as a direct reference that helps the model minimize the risk of semantic drift, where caption expansions may diverge from the intended meaning. 
Our results show that incorporating \textsc{Cece} along with the original caption further improves the image-to-text and text-to-image alignment, providing lexical diversity, improved semantic reasoning, and a more interpretable output, without fine-tuning the models, which may compromise their zero-shot capabilities.

Our contributions are summarized as follows: 
a) We propose Caption Expansion with Contradictions and Entailments (\textsc{Cece}), a principled approach that leverages entailments and contradictions to preserve the semantic meaning while providing lexical diversity of text descriptions.
b) We show that \textsc{Cece} significantly outperforms prior decomposition methods, obtaining $47.5\%$ 
on Winoground (group score) and $47.9\%$
on EqBen (group score) without finetuning.
c) We conduct extensive experiments on benchmarks that score agreement with human judgments of alignment
for image-text alignment.
d) We provide thorough ablation studies and analyses to evaluate the performance of our method under various conditions, and introduce a simple ensembling approach that effectively boosts the accuracy when associating each image-text pair.

\section {Related Work}

\textbf{Single-caption Scoring frameworks.} Commonly used for evaluating the alignment between text and images in VLMs~\citep{instructblip, llava, llavaimproved, llavanext, qwen}, these approaches include similarity scores derived from multimodal encoders \citep{radford2021learning}, as well as text similarity metrics based on image captioning models \citep{li2023blip}. 
However,  summarizing the relationship between text and images using single embeddings often fails to capture the semantic granularity needed for fine-grained image-text alignment~\citep{zhao2024bridging}. Moreover, these metrics are often uncalibrated and may obscure important nuances; for example, a particular CLIPScore value might indicate a good match for pixel art but be considered poor for realistic images, which might affect image-text alignment compared to the human judgments of alignment in text-to-image evaluation metrics. 
More recently, \cite{lin2024vqascore} introduced VQAScore, which leverages a VLM to computes the likelihood of a given image-caption pair, by re-writing the caption as a binary question (``yes$\mid$no"), yielding significant improvements. 
Our approach builds on this while aiming to provide a more detailed and semantically diverse evaluation. By generating entailments and contradictions, we introduce a mechanism to understand the model's nuanced response to positive and negative cues, thereby addressing some limitations of prior single-embedding scoring methods.

\textbf{Sentence Decomposition via Semantics (SDS) frameworks.}
Given the limitations of single-caption scoring frameworks, recent works have explored more sophisticated evaluation methods based on sentence decomposition and semantic analysis~\citep{Khan2023Exploring2310,hu2023tifa,cho2024visual}. These approaches aim to provide a more comprehensive and fine-grained evaluation of text-to-image and image-to-text alignment. Typically, an LLM is instructed to generate subsets of validation questions and expected answers that a VLM can evaluate~\citep{Wan2024CRG}. 
Similarly, \cite{sanders2024tvtreesmultimodalentailmenttrees} uses entailments to improve video question answering.
While these methods provide finer semantic analysis and interpretability to the evaluation process, 
SDS methods typically produce outputs that are direct entailments of the original caption. 
Furthermore, \cite{yarom2024you} introduces VNLI, an approach that finetunes a model that receives an image and a set of entailments and contradictions, where the contradictions are defined as identified question-answer pairs with the lowest VQA score.
We instead directly instruct the LLM to output entailments and contradictions with the input caption as a premise, providing a strong prior for caption expansion and exclusion, without finetuning the models.

\textbf{Chain-of-Thought Prompting frameworks.} 
Recent work has shown promising results when incorporating Chain-of-Thought (CoT) prompting~\citep{wei2022chain} to enhance compositional reasoning in challenging vision-language scenarios, using large-VLMs~\citep{llama, gpt4, hu2024visual-sketchpad}.
Notably, \cite{zhang2024cocotcontrastivechainofthoughtprompting} introduces CoT for multiple image-to-text matching, through a contrastive approach for comparative reasoning.
Similarly, a two-step prompting strategy is introduced to generate descriptions of the given image, which is then used by the model to answer specific questions~\citep{wu2023rolechainofthoughtcomplexvisionlanguage, ossowski2024promptinglargevisionlanguagemodels}.
These works prompt the VLM with multiple images and instruct them to choose the correct one. 
\cite{mitra2024compositional} propose to generate scene graphs as an intermediate reasoning step, and instruct the VLM to pick from two given captions given the image and scene graph. This also reformulates the problem as a multiple-image-to-text matching task.
In contrast, our approach is only allowed to evaluate one image and text at a time -- we argue that this setup closely aligns with real-world scenarios, where compositional understanding must be robust without the benefit of direct comparisons between multiple images or captions. By evaluating each image-caption pair independently, we ensure that the model's reasoning is not influenced by relative comparisons, thus providing a more realistic assessment of its compositional capabilities.

\section{Methodology}
\label{method}

To effectively address the limitations of Vision-Language Models (VLMs) in handling complex compositional visual-textual relationships, we introduce \textsc{Cece}: Caption Expansion with Contradictions and Entailments. Our approach leverages Natural Language Inference (NLI) to systematically generate entailments and contradictions for each image-caption pair, capturing the deeper meaning of the text and a more interpretable metric for image-to-text and text-to-image evaluation and alignment. We describe our proposed method, beginning with the generation of entailments and contradictions via \textsc{Cece} (Section~\ref{sec:framework_step1}). We then explain the likelihood computation for each caption expansion (Section~\ref{sec:framework_step2}). Finally, we describe the score-balancing mechanism we use to integrate the contributions of entailments, contradictions, and the original captions into a unified evaluation framework (Section~\ref{sec:framework_step3}).

\begin{figure}[htbp] 
    \centering
    \includegraphics[width=1.\textwidth]{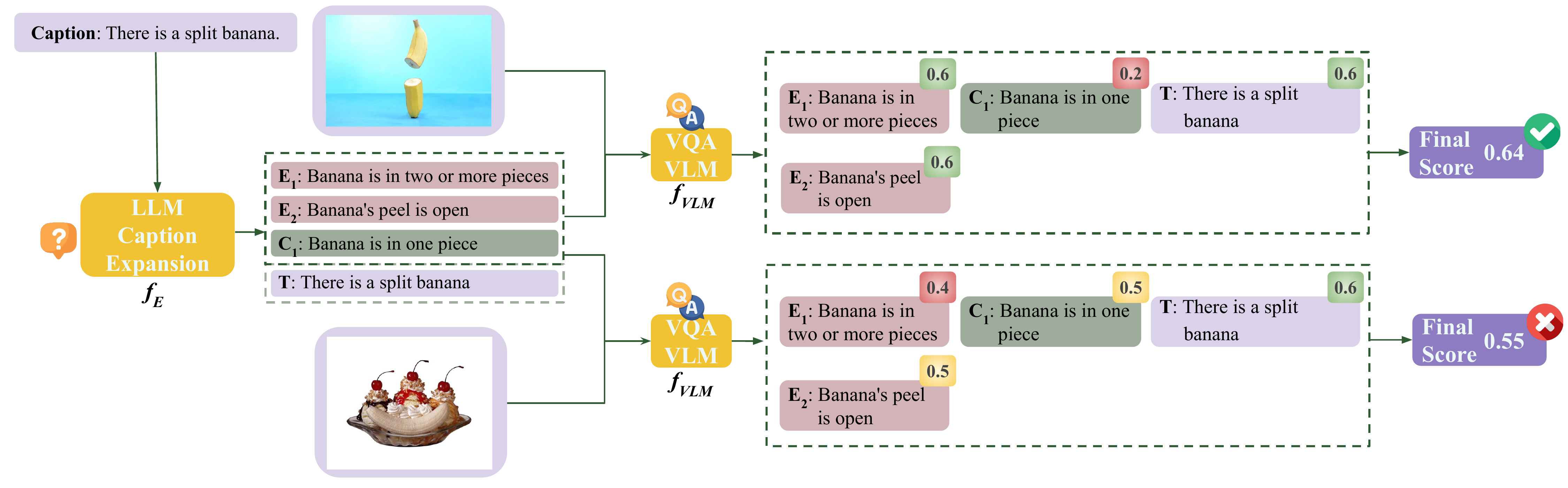} 
    \caption{Complete pipeline of our proposed approach. \textsc{Cece} provides diverse semantic inclusion and exclusion given a caption premise. A VLM is then used to compute the likelihood of the captions generated via entailments and contradictions. The scores are finally balanced along with the original image-text results to avoid semantic drift and enable better alignment.}
    \label{fig:figure2}
\end{figure}

Given an evaluation dataset that consists of image-caption pairs, we define: $X = (T, I)$
as an image-caption pair, where $T$ represents the textual caption and $I$ is the image to evaluate.

\subsection{Caption Expansion} 
\label{sec:framework_step1}
To enrich the semantic representation via caption expansion, we instruct an LLM, denoted as $f_E$, to generate entailments and contradictions (i.e. expansions) based on the original caption $T$ (see Figure~\ref{fig:figure2}). 
This process outputs two subsets: the \textbf{entailment set} $E$, consisting of hypotheses that logically follow from $T$, and the \textbf{contradiction set} $C$, consisting of hypotheses that are logically incompatible with $T$ as follows:
$(E, C) = f_E(T)$. 
Here, $E$ represents the subset of generated entailments: 
    $E = \{e_{1}, e_{2}, \dots, e_{n}\}$,
and $C$ represents the subset of generated contradictions:
    $C = \{c_{1}, c_{2}, \dots, c_{n}\}$.
Each element in $E$ and $C$ aims to provide a diverse but grounded derivation of the original caption through semantic inclusion and exclusion.~\footnote{Prompt details are in Appendix~\ref{app:prompt}.}

\subsection{Likelihood Computation}
\label{sec:framework_step2}
Following~\citet{lin2024vqascore}, we use a Vision-Language Model (VLM), denoted as $f_{\text{VLM}}$, to evaluate image-text pairs. To standardize the input for the VLM, we use a function $q(.)$ that converts a given text $t$ into a question format, which enables the model to assess the visual-textual alignment in a consistent manner. 
Specifically, $q(.)$ generates a yes/no question that captures the essence of the original statement. If we take the example in Figure~\ref{fig:figure2}, the caption $t =$ \textsc{``there is a banana split"} is formatted as $q(t) =$ \textit{``Does \textsc{there is a banana split} can be observed in the image? Answer yes or no"}. 

We proceed to evaluate all captions generated in the previous step as follows:
\begin{compactenum}[(1)]
\item For entailments, we compute the likelihood of answering ``yes" given the image-text pair:
\begin{align}
    f_{\text{VLM}}^{ent}(e_{i}, I) = P(\textit{``yes"} \mid I, q(e_{i}))
\end{align}
\item For contradictions, we compute the likelihood of answering ``no" given the image-text pair:
\begin{align}
    f_{\text{VLM}}^{cnt}(c_{i}, I) = P(\textit{``no"} \mid I, q(c_{i}))
\end{align}
\item For the original caption, we compute the likelihood of answering ``yes" as follows:
\begin{align}
    f_{\text{VLM}}^{cap}(T, I) = P(\textit{``yes"} \mid I, q(T))
\end{align}
\end{compactenum}

This step allows us to assess the model's agreement with both positive cues (entailments) and negative cues (contradictions), as well as its consistency with the original caption. 
In all cases, the probabilities are normalized such that $P(``yes")+P(``no")=1$. By evaluating these three components, we better assess the VLM's ability to align visual content with textual information, capturing nuanced relationships across different semantic variations.

\subsection{Balancing Scores}
\label{sec:framework_step3}
To integrate the information from entailments, contradictions, and the original caption, we employ a two-step balancing process using hyperparameters $\alpha_1$ and $\alpha_2$, to obtain a more comprehensive evaluation of the model's performance while avoiding semantic drift.
By carefully balancing the contributions from entailments, contradictions, and the original caption, this approach ensures that the final assessment remains grounded in the intended meaning of the original text, minimizing the risk of unintended shifts in interpretation that could arise from generated variations (refer to the nuanced scores to the right in Figure~\ref{fig:figure2}).

First, we define the aggregation function $S(.)$, which computes the average score across all elements in a given set. For example, to score all entailment outputs given the set $E$ containing $M$ elements, the aggregation function is given by:
\begin{align} 
    S(E, I) = \frac{1}{M} \sum_{i=1}^{M} f_{\text{VLM}}^{ent}(e_i, I)
\end{align}
and score all contradiction outputs in a similar way:
\begin{align} 
    S(C, I) = \frac{1}{M} \sum_{i=1}^{M} f_{\text{VLM}}^{cnt}(c_i, I)
\end{align}

Then, we compute a weighted sum of the entailment, contradiction, and the original caption score: %
\begin{align}\label{eq:alpha_1}
    S(X) = \underbrace{\alpha_2 \cdot \left[ \alpha_1 \cdot S(E, I) + (1-\alpha_1) \cdot S(C, I) \right]}_{\text{balance the entailments and contradictions}} + \underbrace{(1-\alpha_2) \cdot S(T, I)}_{\text{balance the given premise}}
\end{align} 
where $S(T, I)$ represents the VLM score for the original image-caption pair.

This two-step mechanism allows for a flexible adjustment of the importance assigned to entailments, contradictions, and the original caption in the final assessment, balancing their contributions. By effectively incorporating both positive and negative reasoning cues along with the original context, we aim to achieve a nuanced evaluation that better aligns with human judgments, avoids semantic drifts, and reduces overreliance on biased or superficial features.

\section{Experiment Settings}

\textbf{Baselines.}
We compare \textsc{Cece} with a wide range of baseline methods divided into three categories. 1) End-to-end models (i.e., Single-caption Scoring frameworks) including CLIPScore~\citep{radford2021learning}, BLIP2$_{\text{ITM}}$~\citep{li2023blip}, VQAScore~\citep{lin2024vqascore}, VIEScore~\citep{ku2023viescore}, and GPT4V-Eval~\citep{zhang2023gpt}; 2) Visual Programming (VP) frameworks (VisProg~\citep{gupta2023visual}, ViperGPT~\citep{suris2023vipergpt}, VPEval ~\citep{cho2024visual}); 3) Sentence Decomposition via Semantic (SDS) frameworks (VQ2~\cite{yarom2024you}, DSG~\citep{cho2024visual}, CCoT~\citep{mitra2024compositional}).
Note that VP and SDS frameworks use an LLM for program instruction or sentence decomposition, including  ChatGPT~\citep{ChatGPT}, GPT4~\citep{gpt4}, FlanT5~\citep{chung2024scaling}, and several VLMs, including ViLT~\citep{kim2021vilt}, OWL-ViT~\citep{minderer2022simple}, CLIP~\citep{radford2021learning}, GLIP~\citep{li2022grounded},
GroundDINO~\citep{liu2023grounding},
LLaVA-1.5~\citep{llavaimproved},
LLaVA-1.6~\citep{llavanext},
PaLI-17B~\citep{chen2022pali} and the finetuned model introduced by ~\cite{lin2024vqascore} CLIP-FlanT5-11B.

\textbf{Implementation.} 
We use Llama3.1 70B~\citep{llama} as our LLM for caption expansion through NLI.
We further evaluate \textsc{Cece} on different VLMs (BLIPv2~\citep{li2023blip}, InstructBLIP~\citep{instructblip}, LLaVA-1.5~\citep{llavaimproved}, LLaVA-1.6~\citep{llavanext}) and incorporate a soft-assembling method that balances the results scores of different models, by balancing the scores from entailments and contradictions (VLM scores from LLaVA-1.5), and the original caption (VLM scores from LLaVA-1.6). We use $\alpha_1 = 0.5$ and $\alpha_2 = 0.6$ in all experiments.
Additional details are included in Section~\ref{sec:analysis}.

\textbf{Tasks and Benchmarks.}
We evaluate the compositional capabilities of \textsc{Cece} in three different tasks. 1) Image-text matching evaluation through binary retrieval tasks, which require determining the best caption from a pair of candidates for a given image, as well as determining the best image from a pair of candidates for a given caption. We report results on two benchmarks (Winoground~\citep{thrush_and_ross2022winoground},  EqBen~\citep{wang2023equivariant}) and the performance is evaluated using three metrics: (i) a text score, which assesses the model's ability to identify the correct caption for a given image; (ii) an image score, which measures the model's accuracy in selecting the appropriate image based on a provided caption; and (iii) a group score, which evaluates the successful matching of both pairs. 
2) Score agreement with human judgments of alignment for image-text alignment, using images generated from complex text prompts. We report results on five text-to-image evaluation benchmarks (DrawBench~\citep{saharia2022photorealistic}, EditBench~\citep{wang2023imagen}, COCO-T2I~\citep{lin2014microsoft}, TIFA160~\citep{hu2023tifa}, Pick-a-Pic~\citep{kirstain2023pick}). 3) 3D alignment, which assesses the human judgments of alignment for text-to-3D asset generation. We report results on the StanfordT23D~\citep{wu2024gpt} benchmark with the human ratings collected by \cite{lin2024vqascore}.

\begin{table}[H]
\vspace{-0.2cm}
\caption{Performance on challenging compositional benchmarks that require multi-hop reasoning. \textit{Tools} indicate the vision and language models used for inference. \textit{LLM} indicates the large language model used for generating the visual programming output or sentence decompositions. \textit{DSG$^\dagger$} is the only method that uses a model fine-tuned for this task.}
\vspace{-0.2cm}
\label{table:main_compositional}
\centering
\begin{adjustbox}{max width=\textwidth}
\begin{tabular}{lllccccccc}
\toprule
\multirow{2}{*}{\textbf{Method}} & \multirow{2}{*}{\textbf{Tools-$f_{VLM}$}} & \multirow{2}{*}{\textbf{LLM-$f_E$}} & \multicolumn{3}{c}{\textbf{Winoground}} && \multicolumn{3}{c}{\textbf{EqBen}}\\ 
\cmidrule{4-6} \cmidrule{8-10} 
                &                &              & \textbf{Text} & \textbf{Image} & \textbf{Group} && \textbf{Text} & \textbf{Image} & \textbf{Group}\\ 
\midrule
Random Chance        & -- & -- & 25.0 & 25.0 & 16.7 && 25.0 & 25.0 & 16.7 \\ 
Human Evaluation     & -- & -- & 89.5 & 88.5 & 85.5 && -- & -- & -- \\
\midrule
\rowcolor[HTML]{F2F3F4} 
\multicolumn{3}{l}{\textit{End-to-end models}} \\
CLIPScore~\citep{radford2021learning} & CLIP-L-14  & --  & 27.8 & 11.5 & 7.8 && 35.0 & 35.0 & 25.0\\
BLIP2$_{\text{ITM}}$~\citep{li2023blip} & BLIPv2     & --  & 42.8 & 22.0 & 18.3 && 48.6 & 43.6 & 35.0\\  
VQAScore~\citep{lin2024vqascore} & InstructBLIP & -- & 44.5 & 42.8 & 28.5 && 49.3 & 58.6 & 38.6\\ 
VQAScore~\citep{lin2024vqascore} & LLaVA-1.5 & -- & 45.5 & 41.3 & 29.8 && 45.0 & 47.1 & 28.6\\ 
VQAScore~\citep{lin2024vqascore} & LLaVA-1.6 & -- & 46.8 & 45.8 & 31.3 && 46.4 & 54.3 & 32.9\\ 
VIEScore~\citep{ku2023viescore} & GPT4-Vision     & --  & 40.8  & 39.3  & 34.5 && 40.0 & 34.3 & 32.9\\ 
GPT4V-Eval~\citep{zhang2023gpt} & GPT4-Vision     & --  & 44.5  & 49.0  & 36.3 && 42.9 & 40.0 & 35.0\\ 
\midrule
\rowcolor[HTML]{F2F3F4}
\multicolumn{3}{l}{\textit{Visual Programming (VP)}} \\
VisProg~\citep{gupta2023visual}  & ViLT, OWL-ViT & ChatGPT & 3.5 & 3.5 & 3.5 && 7.9 & 7.9 & 7.9 \\           
ViperGPT~\citep{suris2023vipergpt} & CLIP, BLIP, GLIP & ChatGPT & 7.8 & 7.8 & 7.8 && 4.3 & 4.3 & 4.3\\           
VPEval~\citep{cho2024visual}    & BLIPv2, GroundDINO & ChatGPT & 12.8 & 11.0 & 6.3 && 34.3 & 25.7 & 21.4\\ 
\midrule   
\rowcolor[HTML]{F2F3F4}
\multicolumn{3}{l}{\textit{Sentence Decomposition via Semantics (SDS)}} \\
VQ2~\citep{yarom2024you} & LLaVA-1.5 & FlanT5 & 14.0 & 27.3 & 10.0  && 22.9 & 40.7 & 20.0      \\ 
DSG~\citep{cho2023davidsonian} & LLaVA-1.5 & ChatGPT & 21.0 & 16.8 & 15.5  && 26.4 & 20.0 & 20.0     \\ 
DSG~\citep{cho2023davidsonian} & LLaVA-1.6 & Llama3.1 & 45.8 & 45.8 & 31.3  && 47.1 & 44.3 & 32.1     \\ 
DSG$^\dagger$~\citep{cho2023davidsonian} & CLIP-FlanT5-11B & ChatGPT & 41.0 & 38.3 & 28.3  && 45.7 & 47.9 & 35.0     \\ 
CCoT~\citep{mitra2024compositional} & LLaVA-1.5 & GPT4 & 39.8 & 37.3 & 22.3 && -- & -- & -- \\ 
VQ2~\citep{yarom2024you} & PaLI-17B & FlanT5 & 47.0 & 42.0 & 30.5 && -- & -- & -- \\ 
\midrule
\rowcolor[HTML]{F2F3F4}
\multicolumn{3}{l}{ \textit{Caption Expansion with Contradictions and Entailments (\textsc{Cece})} } \\
\textsc{Cece} (Ours) & BLIPv2       & Llama3.1 & 29.8 & 39.3 & 21.5 && 30.0 & 43.6 & 21.4 \\
\textsc{Cece} (Ours) & InstructBLIP & Llama3.1 & 37.5 & 46.3 & 28.8 && 41.4 & 57.1 & 34.3 \\
\textsc{Cece} (Ours) & LLaVA-1.5    & Llama3.1 & 51.3 & 55.3 & 41.0 && \textbf{58.6} & 57.9 & 41.4 \\
\textsc{Cece} (Ours) & LLaVA-1.6    & Llama3.1 & 52.0 & 61.3 & 42.8 && \textbf{58.6} & 64.3 & 47.1 \\
\textsc{Cece} (Ours)$^{*}$ & LLaVA-1.5, LLaVA-1.6    & Llama3.1 & \textbf{55.0} & \textbf{61.3} & \textbf{47.5} && 57.9 & \textbf{65.0} & \textbf{47.9} \\
\arrayrulecolor[HTML]{000000}
\bottomrule
\end{tabular}
\end{adjustbox}
\end{table}

\section{Results}

\textbf{Image-text matching evaluation through binary retrieval tasks.}
We conduct experiments on Winoground and EqBen. Results are shown in Table~\ref{table:main_compositional}. The entailments and contradictions generated by \textsc{Cece} can be applied to a wide variety of VLMs, this allows for a comprehensive evaluation for a wide range of visual-language model architectures. We demonstrate that our method outperforms prior works (including single-caption scoring methods (i.e, end-to-end models), visual programming, and sentence decomposition approaches that also leverage LLMs). 
For a fair comparison, we took the best SDS method (DSG) and ran their end-to-end framework using Llama3.1 and LLaVA-1.6. We also run DSG with the finetuned method introduced by \cite{lin2024vqascore}.
Note that while \textsc{Cece} outperforms all other methods under similar conditions (one-LLM, one-VLM), the best results are obtained through our score-balancing approach \textsc{Cece}*, which leverages both LLaVA-1.5 (scores from entailments and contradictions) and LLaVA-1.6 (scores from the original caption). 

\begin{table}[H]
\caption{Performance on benchmarks that score agreement with human judgments of alignment for image-text alignment. \textit{Tools} indicate the vision and language models used for inference. \textit{LLM} indicates the large language model to generate the sentence decompositions. Note that none of the models have been specifically fine-tuned for this task.}
\vspace{-0.2cm}
\label{table:genai}
\centering
\begin{adjustbox}{max width=\textwidth}
\begin{tabular}{lllccccc}
\toprule
\textbf{Method} & \textbf{Tools-$f_{VLM}$} & \textbf{LLM-$f_E$} & \textbf{DrawBench} & \textbf{EditBench} & \textbf{COCO-T2I} & \textbf{TIFA160} & \textbf{Pick-a-Pic} \\ 
\midrule
\rowcolor[HTML]{F2F3F4} 
\multicolumn{3}{l}{\textit{End-to-end models}} \\
CLIPScore~\citep{radford2021learning} & CLIP-L-14  & --  & 49.1 & 60.6 & 63.7 & 54.1 & 76.0  \\
BLIP2$_{\text{ITM}}$~\citep{li2023blip} & BLIPv2     & --  & 60.5 & 68.0 & 70.7 & 57.5 & 80.0  \\  
VQAScore~\citep{lin2024vqascore} & InstructBLIP & -- & 82.6 & 75.7 & 83.0 & 70.1 & 83.0  \\ 
VQAScore~\citep{lin2024vqascore} & LLaVA-1.5 & -- & 82.2 & 70.6 & 79.4 & 66.4 & 76.0  \\ 
VIEScore~\citep{ku2023viescore} & GPT4-Vision     & --  & -- & -- & -- & 63.9 & 78.0  \\ 
GPT4V-Eval~\citep{zhang2023gpt} & GPT4-Vision     & --  & -- & -- & -- & 64.0 & 74.0  \\
\midrule   
\rowcolor[HTML]{F2F3F4}
\multicolumn{3}{l}{\textit{Sentence Decomposition via Semantics (SDS)}} \\
VQ2~\citep{yarom2024you} & LLaVA-1.5 & FlanT5 & 52.8 & 52.8 & 47.7 & 48.7 & 73.0  \\ 
DSG~\citep{cho2023davidsonian} & LLaVA-1.5 & ChatGPT & 78.8 & 69.0 & 76.2 & 54.3 & 70.0  \\ 
VQ2~\citep{yarom2024you} & PaLI-17B & FlanT5 & 82.6 & 73.6 & 83.4 & -- & --  \\ 
TIFA~\citep{hu2023tifa} & PaLI-17B & Llama2 & 73.4 & 67.8 & 72.0 & -- & --  \\ 
\midrule
\rowcolor[HTML]{F2F3F4}
\multicolumn{3}{l}{ \textit{Caption Expansion with Contradictions and Entailments (\textsc{Cece})} } \\
\textsc{Cece} (Ours) & LLaVA-1.5    & Llama3.1 &  87.3 & 75.6 & 81.3 & 68.9 & \textbf{86.0} \\
\textsc{Cece} (Ours) & LLaVA-1.6    & Llama3.1 &  86.3 & 75.9 & \textbf{83.8} & \textbf{70.4} & 83.0 \\
\textsc{Cece} (Ours)$^{*}$ & LLaVA-1.5, LLaVA-1.6 & Llama3.1 & \textbf{88.2} & \textbf{76.4} & 83.0 & 69.8 & 85.0 \\
\arrayrulecolor[HTML]{000000}
\bottomrule
\end{tabular}
\end{adjustbox}
\end{table}

\textbf{Score agreement with human judgments of alignment for image-text alignment.}
We show results on five text-to-image evaluation benchmarks in Table~\ref{table:genai}. These results measure the correlation of each method score with human judgments of alignments for an image generated based on a textual prompt. Human ratings are given on a 1-to-5-Likert scale or by assigning a binary match-or-not label. We show AUROC for DrawBench, EditBench, and COCO-T2I, pairwise accuracy for TIFA160, and binary accuracy for Pick-a-Pick. \textsc{Cece} consistently outperforms all prior scoring approaches, indicating that caption expansion via NLI better aligns with human judgments when evaluating text-to-image generation methods.

\begin{table}[H]
\caption{Performance on benchmarks that correlate 3D alignment with human agreement.}
\vspace{-0.2cm}
\label{table:alignment_3d}
\centering
\begin{adjustbox}{width={\textwidth},totalheight={5cm},keepaspectratio} %
\begin{tabular}{l c cc}
        \toprule
        \textbf{Method} & \textbf{Pairwise Acc} & \textbf{Pearson} & \textbf{Kendall} \\
        \midrule
        \rowcolor[HTML]{F2F3F4} 
        \multicolumn{4}{l}{\textit{End-to-end models}} \\
        CLIPScore & 61.0 & 48.1 & 32.6 \\
        BLIPv2Score & 56.6 & 34.3 & 23.4 \\
        InstructBLIP & \textbf{68.0} & 59.5 & \textbf{47.5} \\
        LLaVA-1.5 & 64.9 & 55.8 & 40.8 \\
        \midrule
        \rowcolor[HTML]{F2F3F4} 
        \multicolumn{4}{l}{\textit{Finetuned on human feedback}} \\
        ImageReward & 66.3 & 57.1 & 43.6 \\
        PickScore & 60.1 & 41.3 & 30.3 \\
        HPSv2 & 55.9 & 31.5 & 21.9 \\
        \midrule
        \rowcolor[HTML]{F2F3F4} 
        \multicolumn{4}{l}{ \textit{Caption Expansion with Contradictions and Entailments (\textsc{Cece})} } \\
        w/ LLaVA1.5           & 65.3 & \textbf{57.4} & 41.8 \\
        \bottomrule
    \end{tabular}
\end{adjustbox}
\end{table}

\textbf{3D alignment with human agreement.}
We show results in Table~\ref{table:alignment_3d}. We report the pairwise accuracy along with the Pearson and Kendall coefficients, which assume a linear correspondence between human ratings and metric scores. We follow the setting proposed by \cite{lin2024vqascore} and show that \textsc{Cece} consistently outperforms the base model (LLaVA-1.5).

\section{Analysis}
\label{sec:analysis}

We conduct an in-depth analysis of \textsc{Cece} through multiple perspectives, including detailed breakdowns on Winoground and Eqben benchmarks, lexical diversity, semantic drift, and the impact of incorporating entailments and contradictions. Moreover, we demonstrate the robustness of \textsc{Cece} across different model architectures and present a comprehensive ablation study to understand the importance of each component in our approach.

\textbf{Detailed results on Winoground.}
We show fine-grained results on tags provided by Winoground in Tables~\ref{table:wino_breakdown_ling} and \ref{table:wino_breakdown_vis}. Each sample is grouped per skill category and can include multiple skills. We compare our method against the results from the base end-to-end models, since they outperform prior work based on SDS. Notably, \textsc{Cece} not only outperforms objects and relations from the linguistic side, but it also outperforms in cases where the images need to be interpreted non-literally due to idiomatic uses of language in a caption. It also outperforms the base models when a symbolic description must be understood to make a correct prediction (e.g., typically in non-natural images, such as drawings or illustrations).

\begin{table}[H]
\caption{Detailed analysis on Winoground. Results are grouped by linguistic ($_L$) tags.
We report results using \textsc{Cece} with LLaVA-1.6 and \textsc{Cece}* with LLaVA-1.5 (for entailments and contradictions) and LLaVA-1.6 (for the given caption).
}
\vspace{-0.2cm}
\label{table:wino_breakdown_ling}
\centering
\begin{adjustbox}{width={\textwidth},totalheight={3.5cm},keepaspectratio} %
\begin{tabular}{lccccccccccc}
\toprule
\multirow{2}{*}{\textbf{Method}} & \multicolumn{3}{c}{\textbf{Object$_{L}$}} && \multicolumn{3}{c}{\textbf{Relation$_{L}$}} && \multicolumn{3}{c}{\textbf{Both$_{L}$}} \\ 
\cmidrule{2-4} \cmidrule{6-8}  \cmidrule{10-12} 
 & \textbf{Text} & \textbf{Image} & \textbf{Group} && \textbf{Text} & \textbf{Image} & \textbf{Group} && \textbf{Text} & \textbf{Image} & \textbf{Group} \\
\midrule
Human & 92.20 & 90.78 & 88.65 && 89.27 & 90.56 & 86.70 && 76.92 & 57.69 & 57.69 \\
\midrule
InstructBLIP & 42.5 & 49.7 & 27.7 && 34.3 & 33.9 & 20.2 && \textbf{65.4} & 38.5 & 34.6 \\
LLaVA-1.5 & 43.9 & 48.9 & 32.6 && 52.4 & 42.1 & 33.5 && 53.8 & 30.8 & 26.9 \\
LLaVA-1.6 & 48.2 & 53.9 & 35.5 && 43.8 & 40.8 & 27.9 && \textbf{65.4} & 46.2 & 38.5 \\
\arrayrulecolor[HTML]{D3D3D3}
\midrule
\textsc{Cece} (Ours) & 41.9 & 57.4 & 34.8 && 45.1 & 51.1 & 36.1 && 38.5 & 46.2 & 34.6 \\
\textsc{Cece} (Ours)$^{*}$  & \textbf{56.7} & \textbf{68.8} & \textbf{49.7} && \textbf{53.6} & \textbf{56.7} & \textbf{45.9} && 57.7 & \textbf{61.5} & \textbf{50.0} \\
\arrayrulecolor[HTML]{000000}
\bottomrule
\end{tabular}
\end{adjustbox}
\end{table}

\begin{table}[H]
\caption{Detailed analysis on Winoground. Results are grouped by visual ($_V$) tags.
We report results using \textsc{Cece} with LLaVA-1.6 and \textsc{Cece}* with LLaVA-1.5 (for entailments and contradictions) and LLaVA-1.6 (for the given caption).
}
\vspace{-0.2cm}
\label{table:wino_breakdown_vis}
\centering
\begin{adjustbox}{width={\textwidth},totalheight={3.5cm},keepaspectratio} %
\begin{tabular}{lccccccc}
\toprule
\multirow{2}{*}{\textbf{Method}} & \multicolumn{3}{c}{\textbf{Symbolic$_{V}$}} && \multicolumn{3}{c}{\textbf{Pragmatics$_{V}$}}\\ 
\cmidrule{2-4} \cmidrule{6-8}
 & \textbf{Text} & \textbf{Image} & \textbf{Group} && \textbf{Text} & \textbf{Image} & \textbf{Group}\\
\midrule
Human & 96.43 & 92.86 & 92.86 && 58.82 & 41.18 & 41.18 \\
\midrule
InstructBLIP & 31.7 & 21.9 & 14.6 && 25.0 & 29.2 & 8.3\\
LLaVA-1.5 & 39.0 & 36.6 & 24.4 && \textbf{50.0} & 33.3 & \textbf{33.3}\\
LLaVA-1.6 & 46.3 & 41.5 & 26.8 && 29.2 & 41.7 & 16.7\\
\arrayrulecolor[HTML]{D3D3D3}
\midrule
\textsc{Cece} (Ours) & 51.2 & \textbf{60.1} & \textbf{46.3} && 41.7 & 29.2 & 20.1\\
\textsc{Cece} (Ours)$^{*}$  & \textbf{53.7} & 58.5 & 43.9 && 37.5 & \textbf{41.7} & \textbf{33.3}\\
\arrayrulecolor[HTML]{000000}
\bottomrule
\end{tabular}
\end{adjustbox}
\end{table}

\textbf{On the complexity of image-text matching.}
A key challenge of Winoground is that the captions are also ambiguous; for example, in the sentence \textit{``it hatched before it was eaten"}, \textit{``it"} could refer either to the egg or to the animal inside the egg. 
Previously identified by \cite{winogroundhard}, we show results on the taxonomy of Winoground schemes in  Table~\ref{tab:winohard}. We compare LLaVA-1.6 VQAScore outputs and DSG scores against \textsc{Cece}. For a fair comparison, we report the results of DSG and \textsc{Cece} using Llama3.1 and LLaVA-1.6. 
Notably, DSG outperforms the base model and our method when evaluating samples tagged as \texttt{NonCompositional}. These sample pairs are actually not semantically compositional of one another since they do not contain semantic entities. On the other hand, our \textsc{Cece} shows stronger results for all other tags, where a higher score is expected for the correct image-text pairs when the image is difficult to parse (e.g., objects are small or blurry), the wording of the caption makes it difficult to parse (e.g., \textit{``yellow duck shoes on"}), or common-sense reasoning to match the correct image-text pair is required
(e.g., \textit{``together hammering something"} vs. \textit{``hammering something together"}).
It is important to note that in image-text matching datasets, the captions are syntactically similar, with the key difference of contextual or semantic alterations by swapping objects, relations, or both. 
Similarly, sentences like \textit{``another organism was harmed by a plant, and that plant broke the organism into pieces"} introduce ambiguity regarding the subject, since \textit{"another organism"} could be interpreted as either an animal or possibly another plant.
Results show that \textsc{Cece} is particularly performant also where world knowledge is required.

\begin{table}[H]
\caption{Breakdown analysis with Winoground categories from~\cite{winogroundhard}. For fair comparison, we report numbers for DSG and \textsc{Cece} under similar conditions (i.e., same LLM and VLM).}
\vspace{-0.2cm}
\label{tab:winohard}
\centering
\begin{adjustbox}{width={\textwidth},totalheight={3.5cm},keepaspectratio} %
\begin{tabular}{@{}lccccccccc@{}}
\toprule
 & \multicolumn{3}{c}{\textbf{\textsc{LLaVA-1.6}}} & \multicolumn{3}{c}{\textbf{DSG}} & \multicolumn{3}{c}{\textbf{\textsc{Cece}}} \\ \cmidrule(l){2-4} \cmidrule(l){5-7} \cmidrule(l){8-10} 
 & \textbf{Text} & \textbf{Image} & \textbf{Group} & \textbf{Text} & \textbf{Image} & \textbf{Group} & \textbf{Text} & \textbf{Image} & \textbf{Group} \\ \cmidrule(l){1-2} \cmidrule(l){2-4} \cmidrule(l){5-7} \cmidrule(l){8-10}
\textit{Non Comp.} 
& 60.0 & 50.0 & 40.0 
& \textbf{66.7} & \textbf{53.3} & \textbf{46.7}  
& \textbf{66.7} & \textbf{53.3} & 43.3 \\
\textit{Ambig. Correct} 
& 30.4 & 28.3 & 17.4
& 34.8 & 34.8 & 21.3 
& \textbf{41.3} &\textbf{ 43.5} & \textbf{26.1}  \\
\textit{Visually Difficult} 
& \textbf{47.4} & 34.2 & 23.7
& 26.3 & 28.9 & 15.8 
& 39.5 & \textbf{50.0} & \textbf{28.9}  \\
\textit{Unusual Text} 
& \textbf{40.0} & 32.0 & 26.0 
& 32.0 & 38.0 & 22.0 
& \textbf{40.0} & \textbf{50.0} & \textbf{34.0} \\
\textit{Unusual Image} 
& \textbf{50.0} & 41.1 & \textbf{33.9} 
& 44.6 & 33.9 & 19.6 
& \textbf{50.0} & 58.9 & \textbf{33.9} \\
\textit{Complex Reasoning} 
& \textbf{32.1} & 32.1 & 19.2
& 28.2 & 28.2 & 16.7 
& 29.5 & \textbf{46.2} & \textbf{20.5} \\
\bottomrule
\end{tabular}
\end{adjustbox}
\end{table}

\begin{figure}[H]
    \centering
    \begin{subfigure}{0.33\textwidth}
        \centering
        \includegraphics[width=\textwidth]{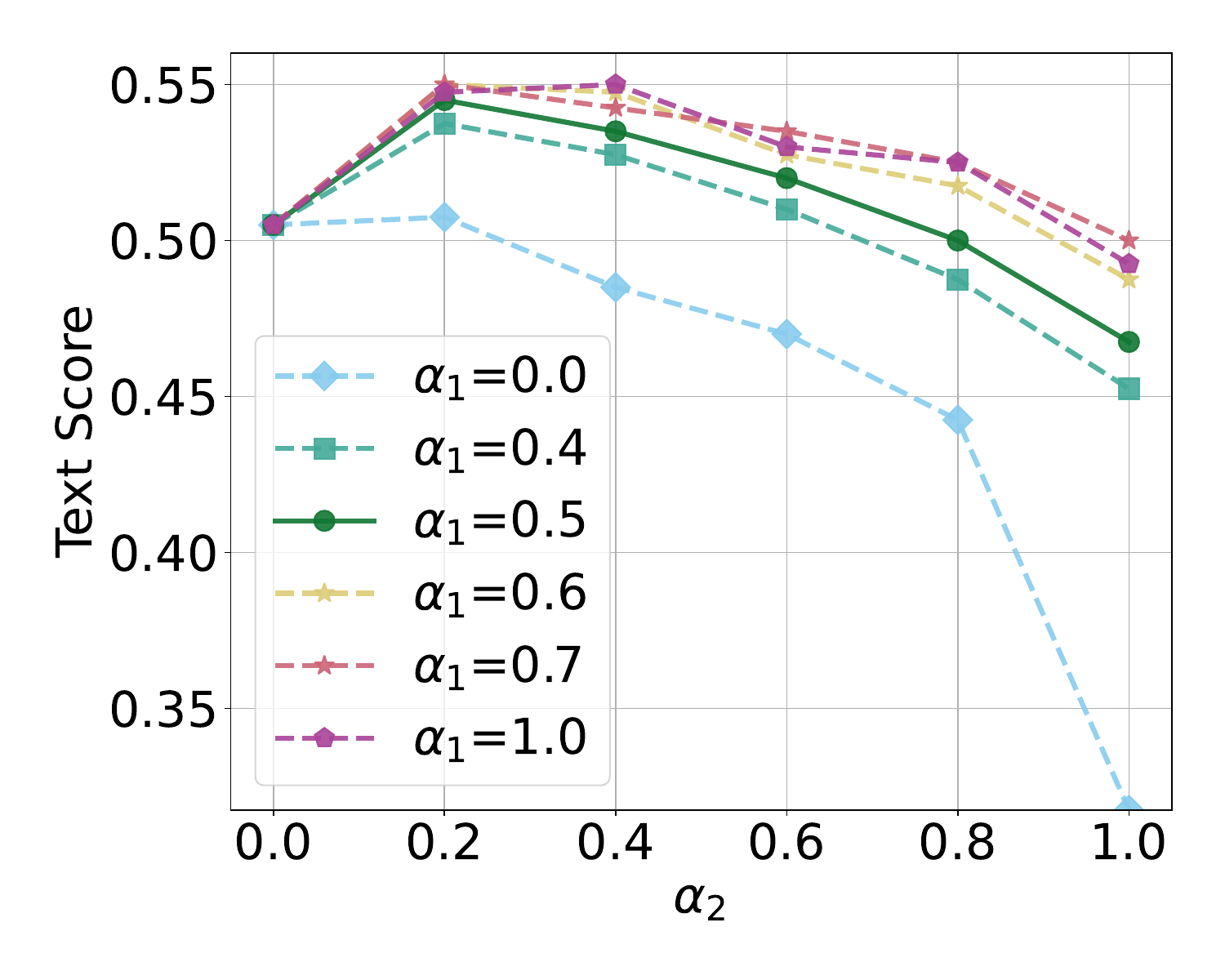}
    \end{subfigure}
    \hfill
    \begin{subfigure}{0.32\textwidth}
        \centering
        \includegraphics[width=\textwidth]{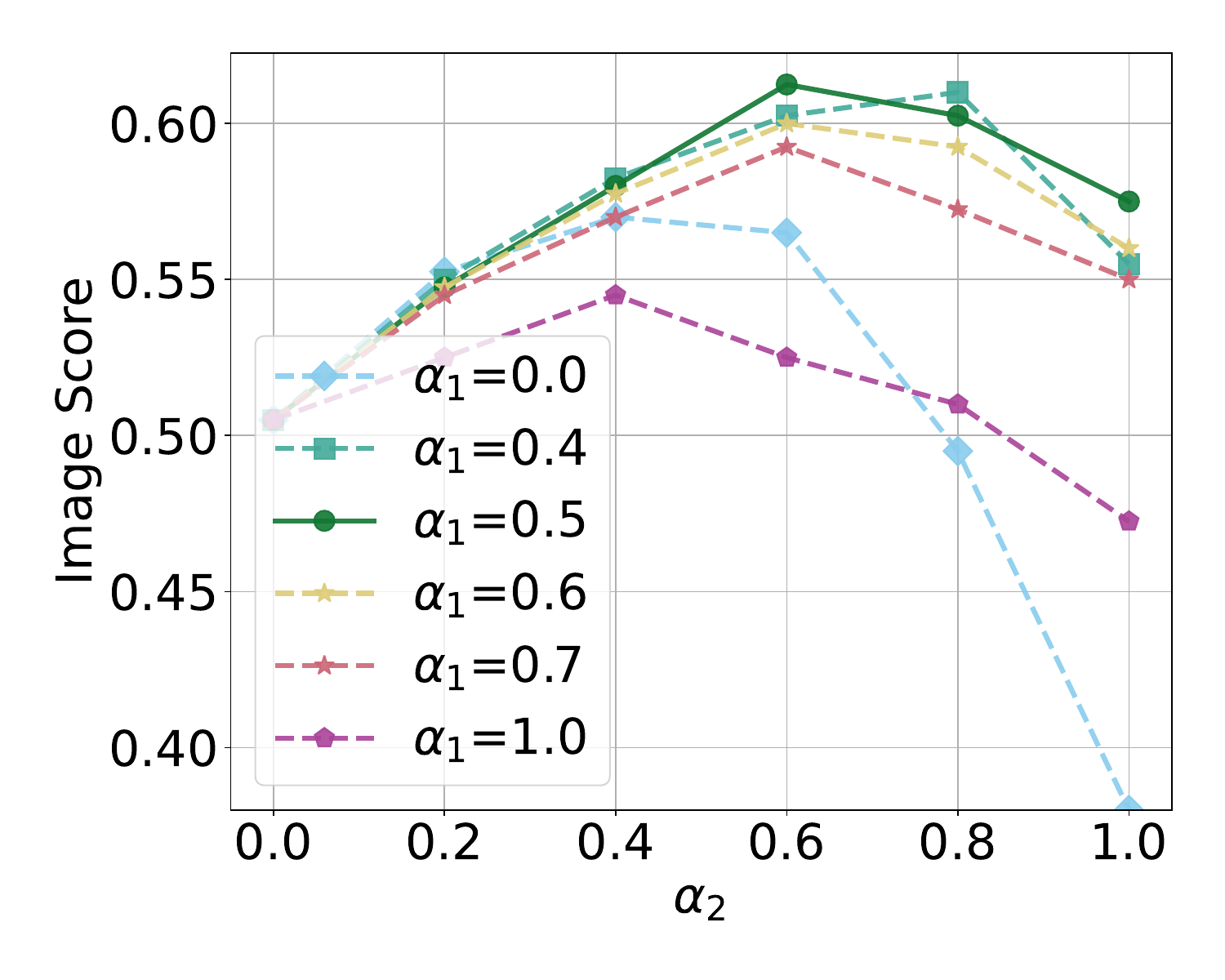}
    \end{subfigure}
    \hfill
    \begin{subfigure}{0.33\textwidth}
        \centering
        \includegraphics[width=\textwidth]{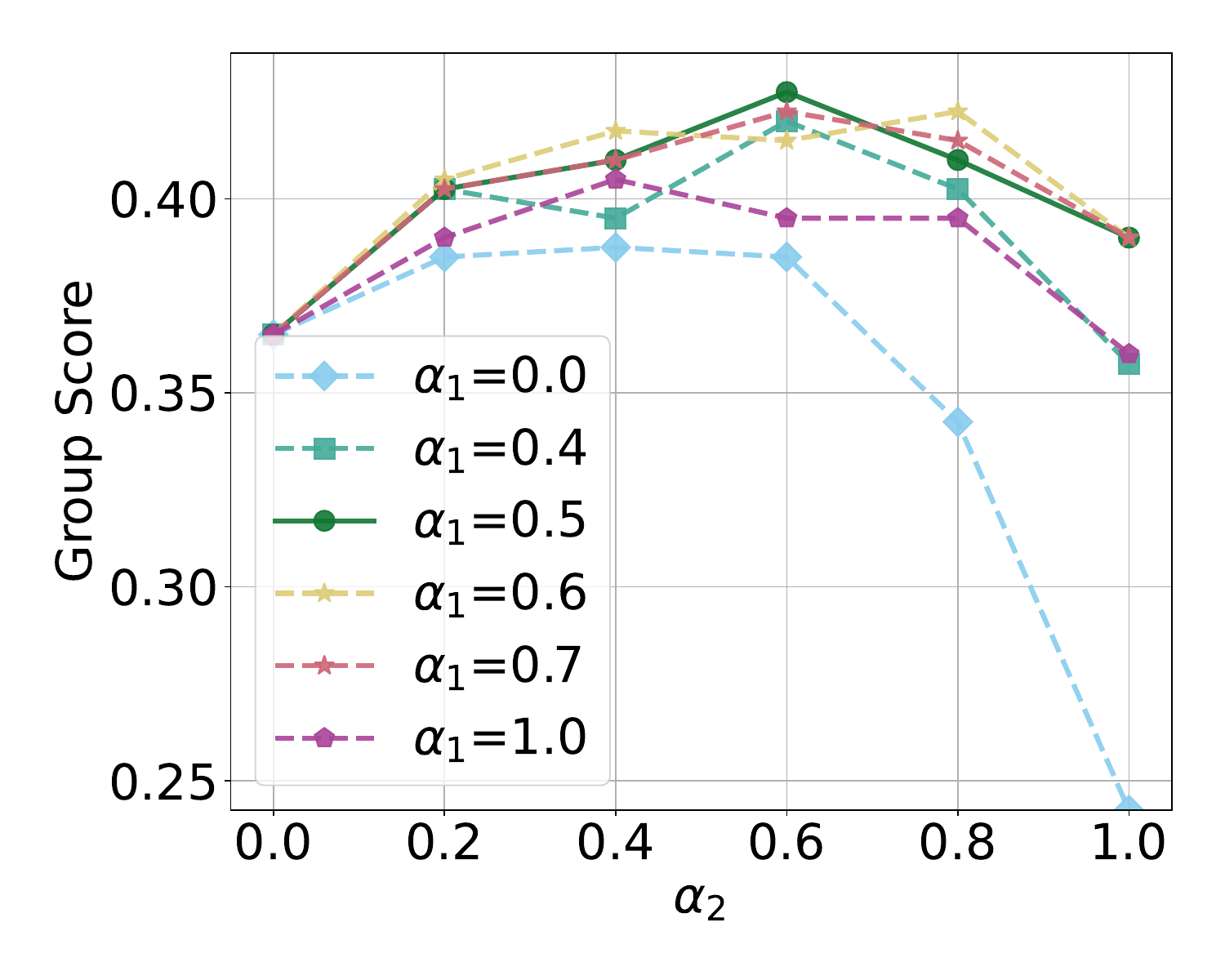}
    \end{subfigure}

    \vspace{-0.2cm}
    
    \begin{subfigure}{0.33\textwidth}
        \centering
        \includegraphics[width=\textwidth]{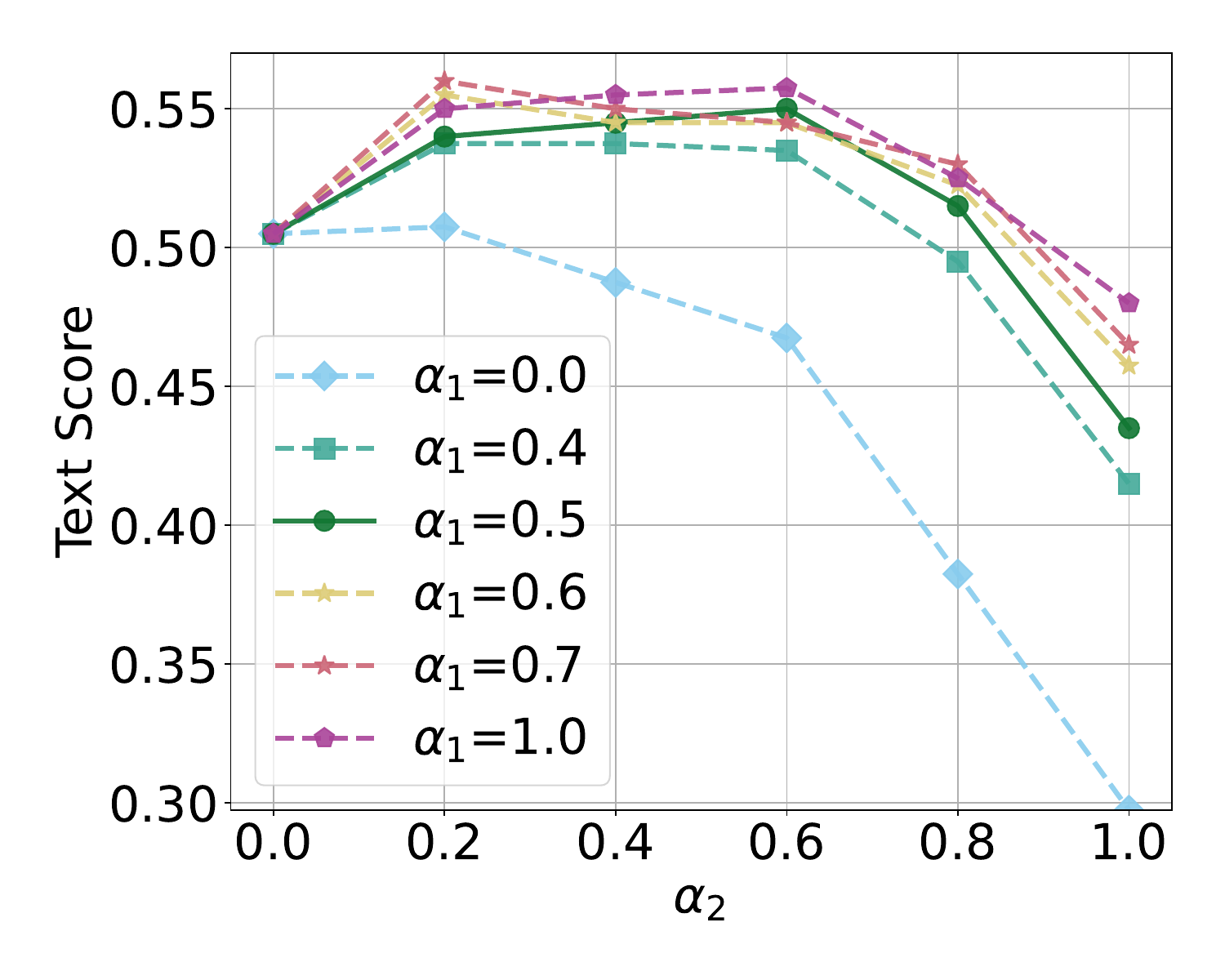}
    \end{subfigure}
    \hfill
    \begin{subfigure}{0.32\textwidth}
        \centering
        \includegraphics[width=\textwidth]{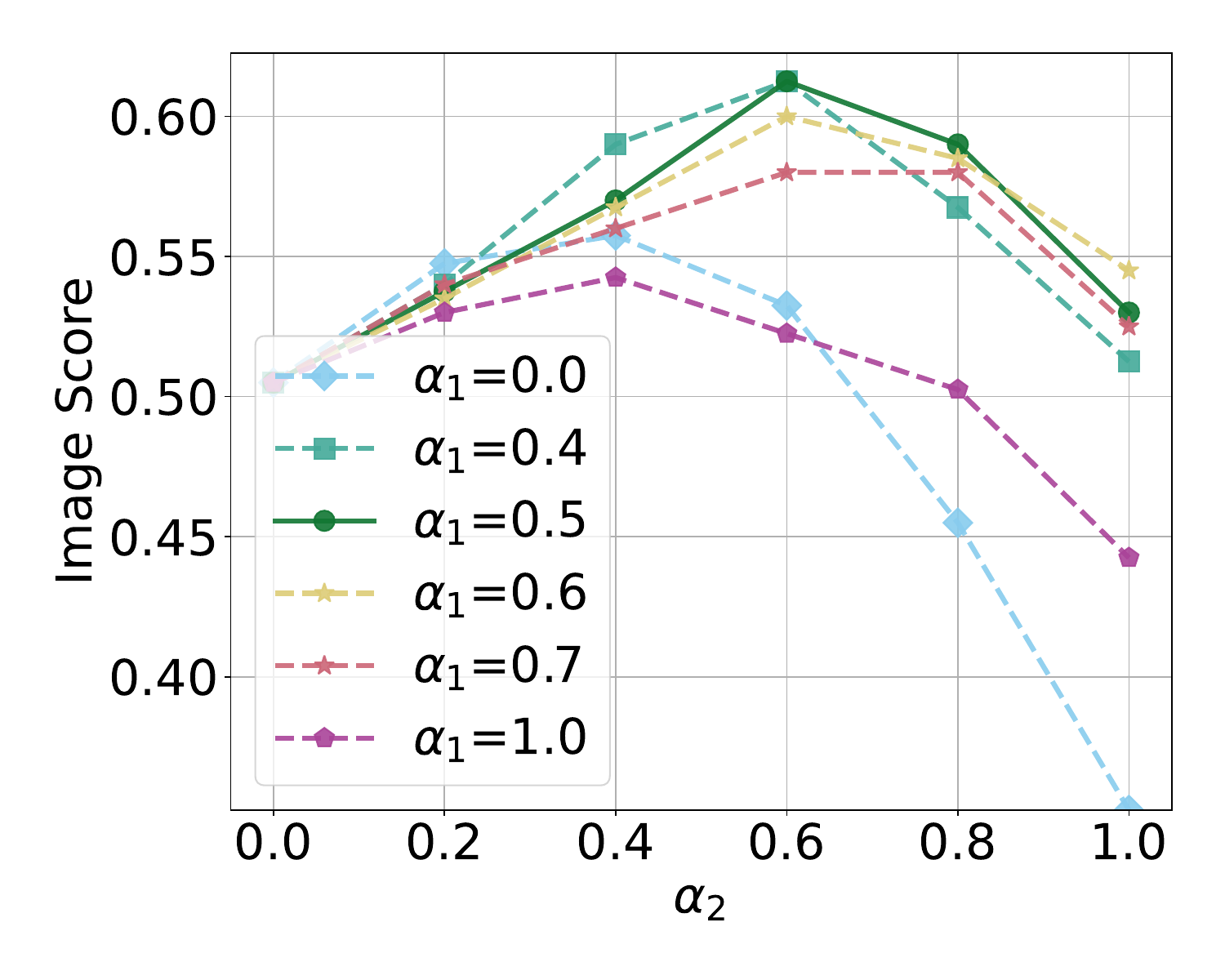}
    \end{subfigure}
    \hfill
    \begin{subfigure}{0.33\textwidth}
        \centering
        \includegraphics[width=\textwidth]{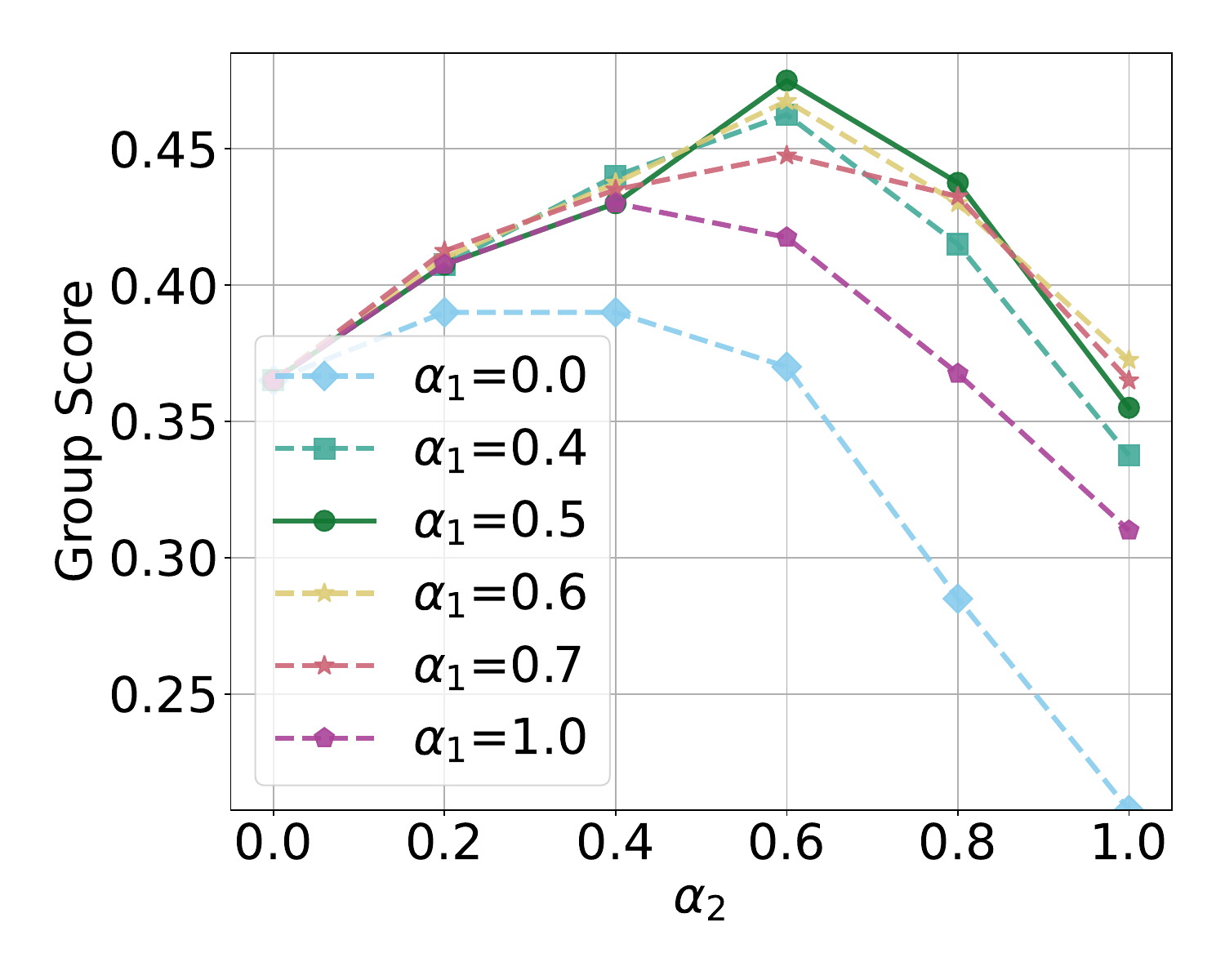}
    \end{subfigure}

    \vspace{-0.3cm}
    \caption{Balancing scores from entailments, contradictions and the original caption. We show how the $\alpha$ values affect the matching performance in Winoground. The first row shows results using LLaVA-1.6 only. The second row shows results using our soft-ensemble approach that balances the scores from LLaVA-1.5 (for entailments and contradictions) and LLaVA-1.6 (for the original caption). Best viewed in color.}
    \label{fig:ablation_alphas}
\end{figure}

\textbf{Semantic drift and balancing scores.}
While the lexical diversity \textsc{Cece} provides benefits the image-text and text-image matching, we also observed a level of divergence from the semantics of the original caption. We refer to this as ``semantic drift", a phenomenon present in LLMs that describes the degradation of text generation quality. \cite{spataru-2024-know} defines this as a degree of separation in generation quality. We mitigate this issue by incorporating the balancing score approach described in subsection~\ref{sec:framework_step3}. We show in  Figure~\ref{fig:ablation_alphas} how different $\alpha$ values balance out the contribution between entailments and contradictions ($\alpha_1$) and the given caption ($\alpha_2$), with a visible trend of highest performance in the middle for each scoring metric. 

\textbf{Lexical diversity and semantic drift.}
We conduct comprehensive experiments to measure the lexical diversity between different semantic decomposition methods (SDS) and \textsc{Cece}. We compute the Jaccard Similarity~\citep{real1996probabilistic} between the Winoground caption and the LLM outputs for each technique. Results show a higher similarity between captions generated via DSG (with a score of $0.53$) in comparison with captions generated via \textsc{Cece} (with a score of $0.32$).

\textbf{Human Validation on Caption Expansion.}
We validate the quality of the entailment and contradiction captions generated by \textsc{Cece} as described in subsection~\ref{sec:framework_step1}. We randomly selected $90$ samples and manually annotated whether the generated captions could be entailed from the original caption. Using a Likert scale ranging from $1$ (definitely not likely) to $5$ (definitely likely), the entailment captions received an average score of $4.7$, while the contradiction captions received an average score of $1.7$. This indicates that the entailment and contradictions are both accurately generated.

\textbf{Importance of each component and soft-ensembling.}
We ablate each component of our proposed method in three ways. We show in Table~\ref{table:ablation_components} results on Winoground and Eqben when using (i) only entailments, (ii) entailments and contradictions, (ii) entailments, contradictions and the given caption. Our results show that entailments alone outperform prior methods, while all components progressively boost the final matching performance.
We further explored different combinations to mitigate the ``semantic drift" problem by balancing the matching scores from different models. We show in Table~\ref{table:ablation_ensembling} how combining different models boosts the final compositional evaluation. 

\begin{table}[H]
\caption{Importance of each component. While entailments alone significantly improve the compositional scoring performance, progressively adding each proposed component yields the best matching score. Results with Llama3.1 and LLaVA-1.6.}
\vspace{-0.1cm}
\label{table:ablation_components}
\centering
\begin{adjustbox}{width={\textwidth},totalheight={2.5cm},keepaspectratio}
\begin{tabular}{cccccccccc}
\toprule
\multirow{2}{*}{\textbf{Entail.}} & \multirow{2}{*}{\textbf{Contrad.}} & \multirow{2}{*}{\textbf{Caption}} & \multicolumn{3}{c}{\textbf{Winoground}} && \multicolumn{3}{c}{\textbf{EqBen}}\\ 
\cmidrule{4-6} \cmidrule{8-10} 
                &                &              & \textbf{Text} & \textbf{Image} & \textbf{Group} && \textbf{Text} & \textbf{Image} & \textbf{Group}\\ 
\midrule
\cmark\ &  &  &  49.3 & 47.3 & 36.0 && 45.0 & 57.9 & 33.6 \\
\cmark\ & \cmark\ &  &  46.8 & 57.5 & 39.0 && 47.1 & 60.0 & 38.6 \\
\cmark\ & \cmark\ & \cmark\ &  52.0 & 61.3 & 42.8 && 58.6 & 64.3 & 47.1 \\         
\bottomrule
\end{tabular}
\end{adjustbox}
\vspace{-0.2cm}
\end{table}

\begin{table}[H]
\caption{Mixture of VLMs. We show results of balancing the scores of entailments and contradictions ($\alpha_1$) and the given caption ($\alpha_2$) with different models.}
\vspace{-0.2cm}
\label{table:ablation_ensembling}
\centering
\begin{adjustbox}{width={\textwidth},totalheight={2.5cm},keepaspectratio} %
\begin{tabular}{cccccccccc}
\toprule
\multirow{2}{*}{\textbf{InstructBLIP}} & \multirow{2}{*}{\textbf{LLaVA-1.5}} & \multirow{2}{*}{\textbf{LLaVA-1.6}} & \multicolumn{3}{c}{\textbf{Winoground}} && \multicolumn{3}{c}{\textbf{EqBen}}\\ 
\cmidrule{4-6} \cmidrule{8-10} 
                &                &              & \textbf{Text} & \textbf{Image} & \textbf{Group} && \textbf{Text} & \textbf{Image} & \textbf{Group}\\ 
\midrule
\cmark\ & \cmark\ &  &  49.3 & 55.5 & 41.0 && 48.6 & 58.6 & 40.7 \\        
\cmark\ &  & \cmark\ &  51.5 & 57.3 & 42.5 && 51.4 & 57.9 & 40.0 \\
 & \cmark\ & \cmark\ &  55.0 & 61.3 & 47.5 && 57.9 & 65.0 & 47.9 \\
\bottomrule
\end{tabular}
\end{adjustbox}
\vspace{-0.2cm}
\end{table}

\section{Conclusion}

In this work, we introduce Caption Expansion with Contradictions and Entailments (\textsc{Cece}), a principled approach designed to enhance compositional reasoning in vision-language models. \textsc{Cece} leverages Natural Language Inference to generate diverse entailments and contradictions, aimed to expand the semantic understanding of textual descriptions. We conduct extensive evaluations across multiple compositional benchmarks, including Winoground and EqBen, and demonstrate that \textsc{Cece} significantly outperforms prior methods without requiring additional fine-tuning, achieving notable results in alignment with human judgments for text-to-image evaluation. Through comprehensive analysis, we show that \textsc{Cece} enhances interpretability and provides a balanced semantic representation, which is crucial for nuanced image-text matching and reasoning. Our results indicate that combining both entailments and contradictions allows vision-language models to consider both inclusive and exclusive semantic cues, leading to interpretable and less biased compositional reasoning. We encourage future work on interpretable multimodal frameworks that can leverage structured semantic expansions across diverse domains and tasks.

\textbf{Broader Impact.}
\textsc{Cece} effectively improves the interpretability and robustness of vision-language models, and contributes to fairer AI systems that align more closely with human reasoning, reducing its reliance on superficial correlations. \textsc{Cece} also has the potential to improve a wide range of applications, such as assistive technologies for people with visual impairments, educational tools, and creative content generation. However, like other works that leverage LLMs, \textsc{Cece} also poses risks related to the generation of misleading content or misuse in malicious contexts. The enhanced contextual interpretation and semantic descriptions could be used to make fabricated or altered visual content more convincing, amplifying the risks associated with misinformation. We encourage the responsible use of frameworks like \textsc{Cece}, with an emphasis on transparency, ethical guidelines, and mechanisms for monitoring and mitigating potential misuse.

\textbf{Acknowledgments.}
Rachel Rudinger is supported by NSF CAREER Award No. 2339746.

\bibliography{iclr2025_conference}
\bibliographystyle{iclr2025_conference}

\clearpage
\appendix
\section{Appendix}

\subsection{Fine-tuned Models}
We also compare our method when using models finetuned with targeted data. In Table~\ref{table:appendix_compositional}, we show results of \textsc{Cece} when using the best performant end-to-end finetuned model (i.e., CLIP-FlanT5-11B~\citep{lin2024vqascore}). Notably, our method improves the results while enhancing the interpretability by including entailments and contradictions in the scoring process.

\subsection{Prompt Details}
\label{app:prompt}
In the caption expansion step, we instruct the LLM to generate both entailments and contradictions in a single output rather than in separate steps, as this helps the LLM maintain a balanced semantic context, leading to more coherent and complementary outputs.
Since the LLM often tends to output simple negative statements for contradictions (e.g., using words like \textit{``no"} or \textit{``not"}), we explicitly instruct the model to avoid such terms. This ensures that the generated contradictions are more diverse and meaningful, rather than being straightforward negations. In addition, following prior work~\citep{mitra2024compositional}, we instruct the LLM to output the generated entailments and contradictions in \textit{JSON}, as it follows an easy format to parse the outputs.
Prompt template is in Figure~\ref{fig:prompt}.

\subsection{Error Analysis}
\label{app:error_analysis}
We further examine cases where \textsc{Cece} fails comparing with using original caption only (LLaVA-1.6) and Sentence Decomposition via Semantics (SDS).
We show detailed entailments and contradictions in \textsc{Cece} column, where \textbf{\textsc{Cece} Final Score} is from Equation~\ref{eq:alpha_1} with $\alpha_1 = 0.5$ and \textbf{Overall Final Score} is from Equation~\ref{eq:alpha_1} where $\alpha_2 = 0.6$. Note that both $\alpha$ values are kept consistent throughout all our experiments.

Figure~\ref{fig:case_1} shows examples where LLaVA-1.6 correctly predicts the score relationship between images given the text but both SDS and \textsc{Cece} fail.
Figure~\ref{fig:case_2} shows examples where LLaVA-1.6 and SDS correctly predict the score relationship between images given the text but \textsc{Cece} fails.
Figure~\ref{fig:case_3} shows examples where LLaVA-1.6, SDS and \textsc{Cece} all fail.

~\\

\begin{table}[H]
\caption{Results on fine-tuned models for compositional benchmarks that require multi-hop reasoning. \textit{Tools} indicate the vision and language models used for inference. \textit{LLM} indicates the large language model used for generating the visual programming output or sentence decompositions.}
\label{table:appendix_compositional}
\centering
\begin{adjustbox}{max width=\textwidth}
\begin{tabular}{lllccccccc}
\toprule
\multirow{2}{*}{\textbf{Method}} & \multirow{2}{*}{\textbf{Tools-$f_{VLM}$}} & \multirow{2}{*}{\textbf{LLM-$f_E$}} & \multicolumn{3}{c}{\textbf{Winoground}} && \multicolumn{3}{c}{\textbf{EqBen}}\\ 
\cmidrule{4-6} \cmidrule{8-10} 
                &                &              & \textbf{Text} & \textbf{Image} & \textbf{Group} && \textbf{Text} & \textbf{Image} & \textbf{Group}\\ 
\midrule
Random Chance        & -- & -- & 25.0 & 25.0 & 16.7 && 25.0 & 25.0 & 16.7 \\ 
Human Evaluation     & -- & -- & 89.5 & 88.5 & 85.5 && -- & -- & -- \\
\midrule
\rowcolor[HTML]{F2F3F4} 
\multicolumn{3}{l}{\textit{Finetuned end-to-end models}} \\
PickScore~\citep{kirstain2023pick} & CLIP-H-14 & -- & 23.8 & 12.5 & 6.8 && 35.7 & 39.3 & 23.6 \\
ImageReward~\citep{xu2024imagereward} & BLIPv2 & -- & 42.8 & 15.3 & 12.8 && 37.9 & 36.4 & 26.4 \\
HPSv2~\citep{wu2023human} & CLIP-H-14 & -- & 11.5 & 7.8 & 4.0 && 27.9 & 26.4 & 17.1 \\
VQAScore~\citep{lin2024vqascore} & CLIP-FlanT5-11B & -- & \textbf{60.3} & 56.3 & 45.3 && 59.3 & 63.6 & 47.9\\ 
\midrule
\rowcolor[HTML]{F2F3F4}
\multicolumn{3}{l}{\textit{Sentence Decomposition via Semantics (SDS)}} \\
DSG$^\dagger$~\citep{cho2023davidsonian} & CLIP-FlanT5-11B & ChatGPT & 41.0 & 38.3 & 28.3  && 45.7 & 47.9 & 35.0     \\ 
\midrule
\rowcolor[HTML]{F2F3F4}
\multicolumn{3}{l}{ \textit{Caption Expansion with Contradictions and Entailments (\textsc{Cece})} } \\
\textsc{Cece} (Ours)$^{**}$ & CLIP-FlanT5-11B & Llama3.1 & 58.0 & \textbf{61.3} & \textbf{48.3} && \textbf{60.0} & \textbf{66.4} & \textbf{49.3} \\
\bottomrule
\end{tabular}
\end{adjustbox}
\end{table}

\begin{figure}[t]
\begin{prompt}[title={Prompt Template}, label=prompt:full_prompt]
\texttt{Given the sentence: \textbf{\{Caption\}} \\
Think step by step. What could be entailed? \\ \\
First, provide two concise descriptions that entail the sentence. Include attributes mentioned in the sentence (color, size, position, amounts). If there is a verb, rephrase it to entail the sentence. \\
Some possible verb entailments include: \\
looking -> (participant) person's eyes over something (do not include the recipient) \\
kissing -> (participant) person's lips touching someone (do not include the recipient) \\
talking -> (participant) person's mouth open (do not include the recipient) \\
hugging -> (participant) person's arms reaching the other person (recipient) \\
person hitting -> (participant) person is in motion \\
object hitting -> (participant) object is damaged \\ \\
The descriptions need to be specific and implicitly entailed in the sentence. Include world knowledge and common sense assumptions. \\
For example: \\
Sentence: A yellow unicorn talks to a tall person \\
Nouns: plant, unicorn, ball, person, sky \\
Entailed descriptions: [Yellow unicorn's mouth is open, Yellow unicorn is gesturing] \\ \\
Then, provide the opposite sentence to each sentence, do not include negations. \\
For example: \\
Sentence: Person talks to unicorn \\
Nouns: unicorn, ball, person, sky \\
Opposite descriptions: [Tall person's mouth is open, Tall person is gesturing] \\ \\
Finally, output the entailed descriptions in a json format. \\
\{ \\
"Entailed descriptions": [Yellow unicorn's mouth is open, Yellow unicorn is gesturing] \\
"Opposite descriptions": [Tall person's mouth is open, Tall person is gesturing] \\
\}
}
\end{prompt}
\caption{Prompt template used in Caption Expansion. \texttt{\textbf{\{Caption\}}} is replaced for each sample.}
\label{fig:prompt}
\end{figure}

\begin{figure}[t]
\centering
    \includegraphics[width=\textwidth]{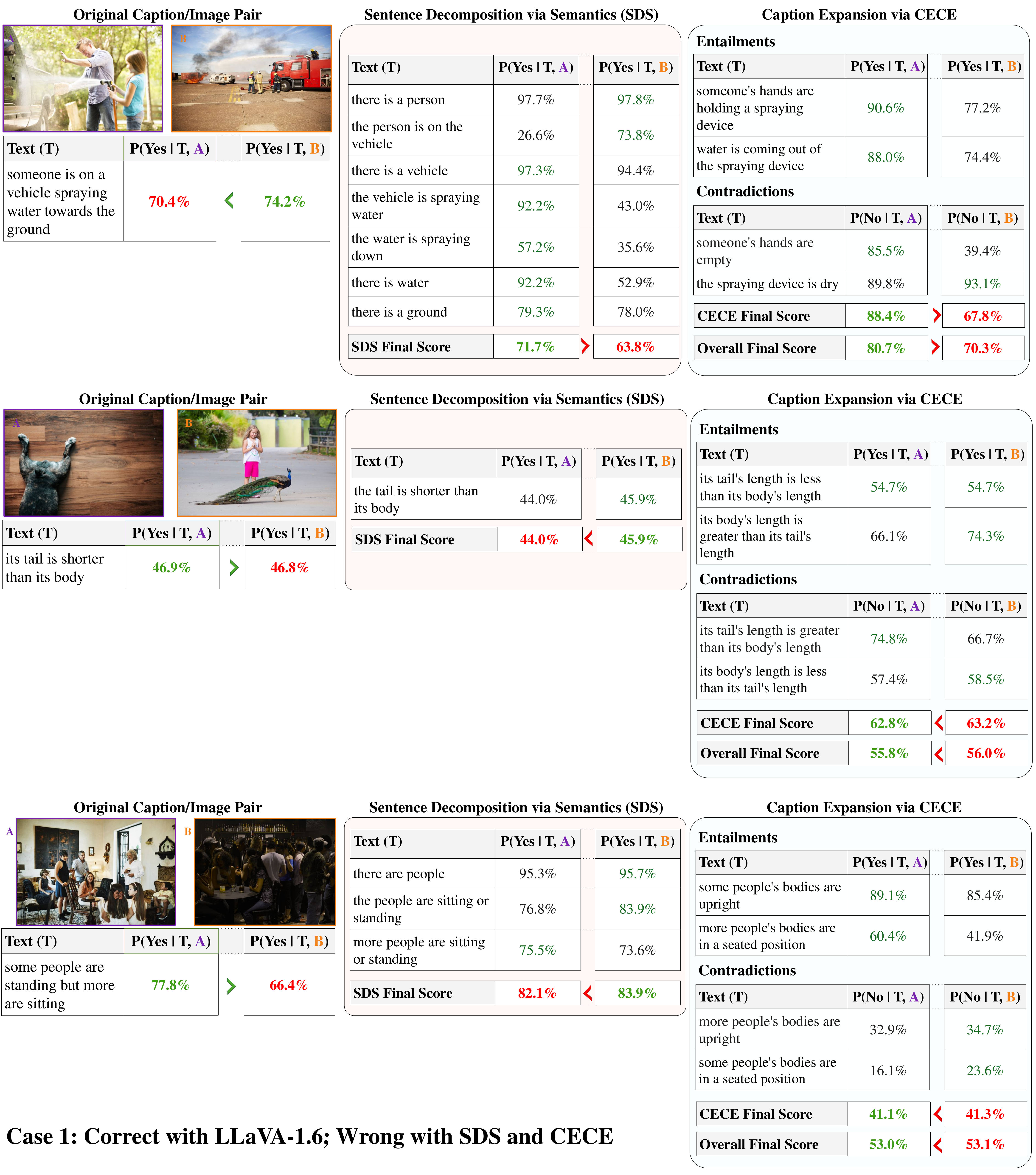} 
    \vspace{0.1cm}
    \caption{Qualitative error analysis: cases where both SDS and \textsc{Cece} fail: 
    a) In the first case, SDS correctly decomposes the caption and focuses on the \textit{vehicle}, which is the one \textit{spraying water}. On the other hand, through \textsc{Cece}, the LLM incorrectly focuses on a person \textit{spraying water}. However, both cases fail since the VLM is unable to identify the vehicle \textit{spraying water}. 
    b) In the second case, SDS fails to decompose the caption into smaller parts, and repeats the given text. Although \textsc{Cece} produces correct entailments and contradictions, the VLM fails to match the correct image-text pair. 
    c) Similarly, for the third case, the VLM seems to fail to match the correct image, which seems too difficult to parse due to out-of-focus and illumination issues.
    }
\label{fig:case_1}
\end{figure}

\begin{figure}[t]
\centering
    \includegraphics[width=\textwidth]{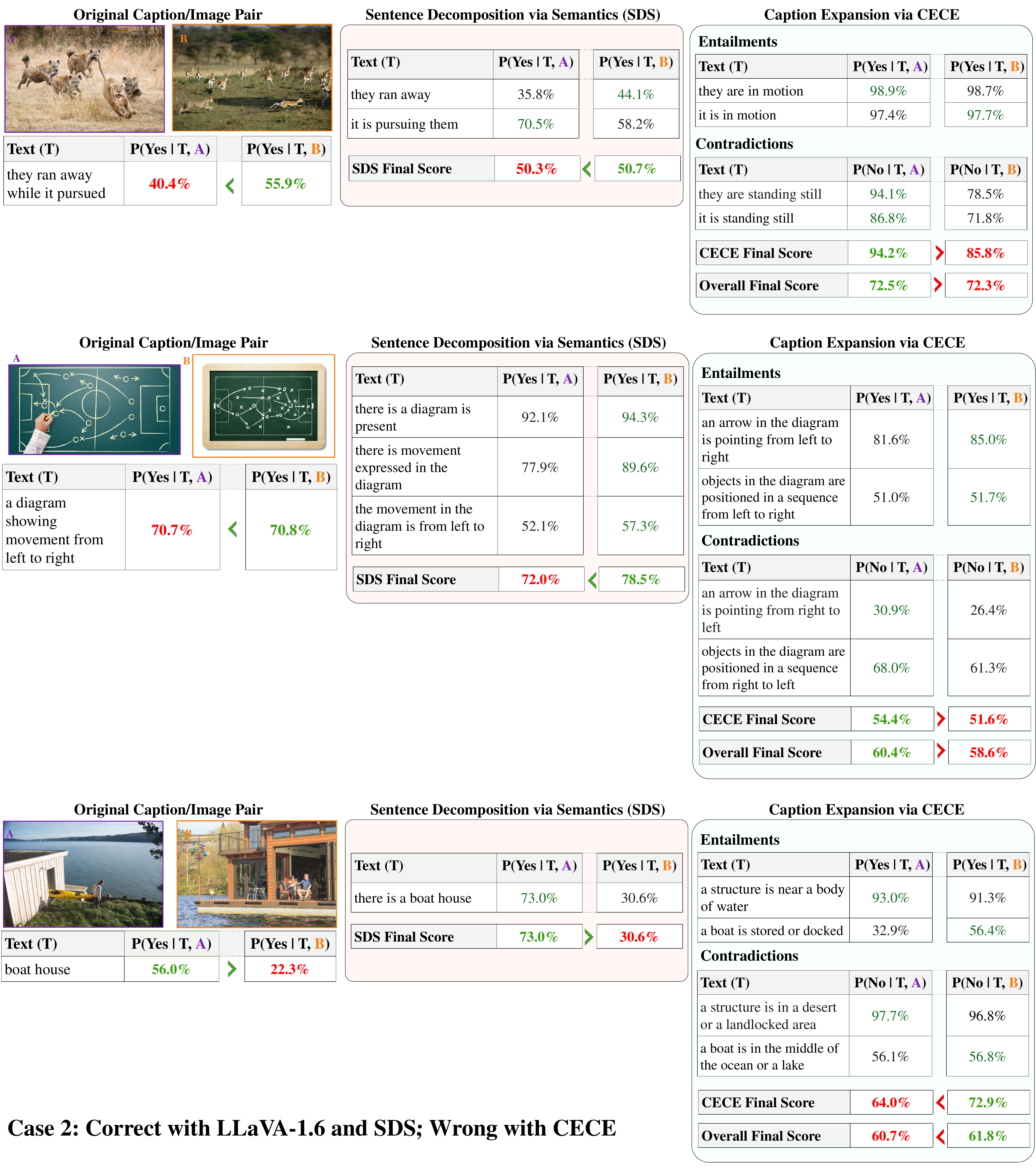} 
    \vspace{0.1cm}
    \caption{Qualitative error analysis: cases where only \textsc{Cece} fails:
    a) In the first case, SDS correctly decomposes the caption, although \textit{they} and \textit{it} are not concretely stated (e.g., could refer to animals or objects), breaking the sentence into two separate statements is sufficient for the VLM to match the correct image. On the other hand, the contradictions generated via \textsc{Cece} introduce statements that are correct, but lead to incorrect conclusions given the nature of the data (i.e., still images).
    b) Similarly, the contradictions generated for the second case introduce incorrect negative statements that weigh over the entailments and lead to the incorrect matching.
    c) In the third case, SDS is unable to decompose the sentence, and the given description contains the name of the referring object (i.e., \textit{boat house}). Although both entailments and contradictions generated via \textsc{Cece} provide a more detailed or fine-grained set of descriptions, the VLM is unable to identify the correct pair.
    }
\label{fig:case_2}
\end{figure}

\begin{figure}[t]
\centering
    \includegraphics[width=\textwidth]{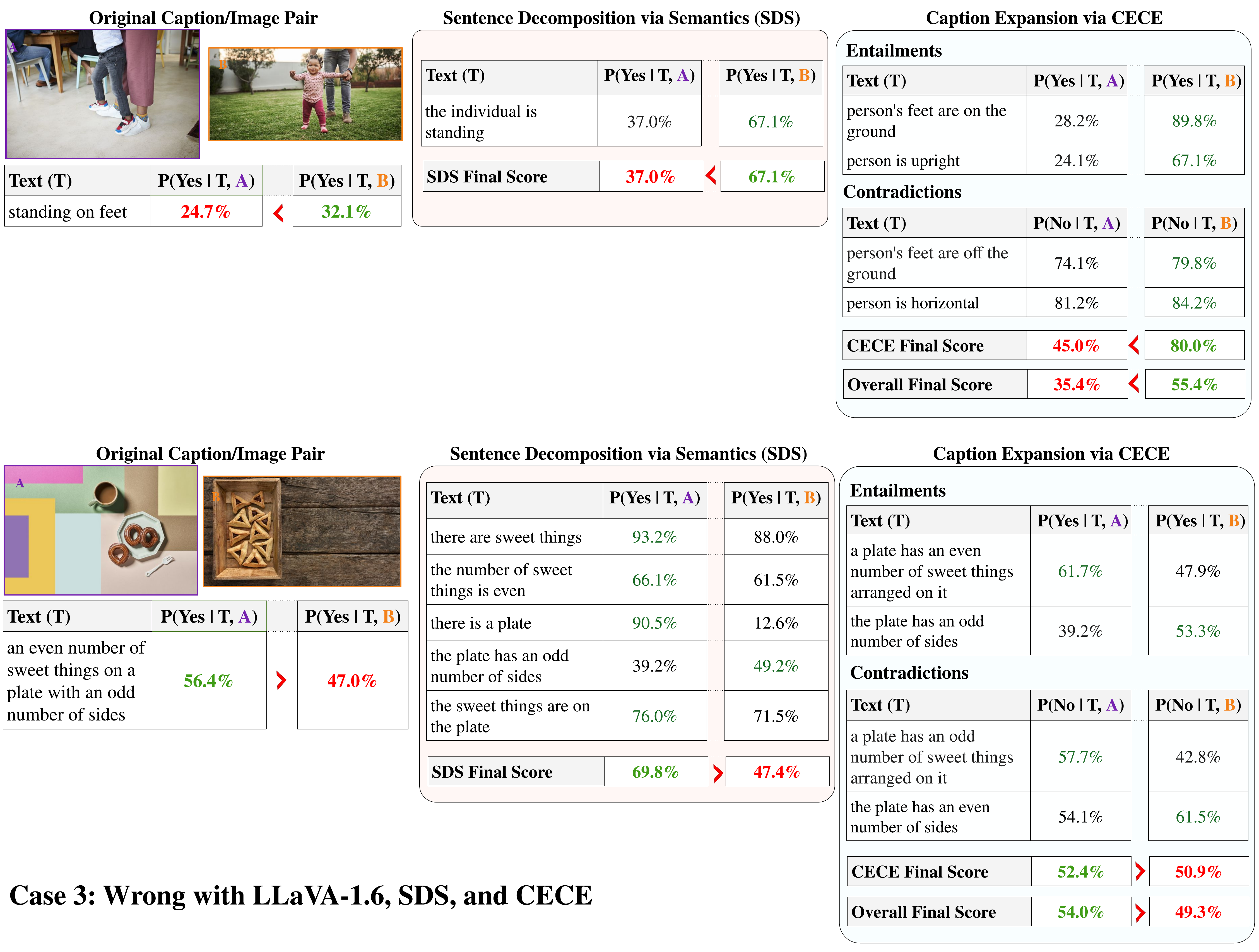} 
    \vspace{0.1cm}
    \caption{Qualitative error analysis: cases where LLaVA-1.6, SDS and \textsc{Cece} all fail:
    a) In the first case, while \textsc{Cece} correctly generates entailments and contradictions, the VLM is unable to match the correct image-text pair. In this case, both images are cutout and it may be difficult for the model to identify the people in the scene. Note that SDS is unable to break the sentence.
    b) As opposed to the previous one, in the second case, SDS is able to correctly decompose the given caption, but the VLM is unable to score the matching image-text pair. \textsc{Cece} on the other hand, fails to generate meaningful entailments and contradictions, leading to an incorrect output. However, the failure margin from \textsc{Cece} is lower than that produced by SDS.
    }
\label{fig:case_3}
\end{figure}

\end{document}